\documentclass[10pt,twocolumn,letterpaper]{article}

\usepackage{cvpr}              
\definecolor{cvprblue}{rgb}{0.21,0.49,0.74}
\usepackage[pagebackref,breaklinks,colorlinks,allcolors=cvprblue]{hyperref}

\usepackage{amsmath}
\usepackage{amsfonts}
\usepackage{multirow}
\usepackage{amssymb}
\usepackage{pifont}
\usepackage{tcolorbox}
\usepackage{listings}

\def\paperID{10523} 
\def\confName{CVPR}
\def\confYear{2026}

\title{TrajRAG: Retrieving Geometric-Semantic Experience for Zero-Shot Object Navigation}

\author{Yiyao Wang$^{1,2}$, Sixian Zhang$^{1,2}$, Keming Zhang$^{1,2}$, Xinhang Song$^{1,2}$\footnotemark[1], Songjie Du$^{2}$, Shuqiang Jiang$^{2,3}$\\
\small \textsuperscript{1}State Key Laboratory of AI Safety, Institute of Computing Technology, Chinese Academy of Sciences, Beijing\\  
\small \textsuperscript{2}University of Chinese Academy of Sciences, Beijing,
\small \textsuperscript{3}Institute of Computing Technology, Chinese Academy of Sciences, Beijing
\\
\tt\small \{yiyao.wang, sixian.zhang, keming.zhang, xinhang.song\}@vipl.ict.ac.cn, \\ 
\tt\small dusongjie25@mails.ucas.ac.cn, sqjiang@ict.ac.cn\\
}

\begin{document}
\maketitle
\renewcommand{\thefootnote}{\fnsymbol{footnote}}
\footnotetext[1]{Corresponding author.}
\begin{abstract}
Existing zero-shot Object Goal Navigation (ObjectNav) methods often exploit commonsense knowledge from large language or vision-language models to guide navigation. 
However, such knowledge arises from internet-scale text rather than embodied 3D experience, and episodic observations collected during navigation are typically discarded, preventing the accumulation of lifelong experience.
To this end, we propose Trajectory RAG (TrajRAG), a retrieval-augmented generation framework that enhances large-model reasoning by retrieving geometric–semantic experiences.  
TrajRAG incrementally accumulates episodic observations from past navigation episodes. 
To structure these observations, we propose a topological-polar (topo-polar) trajectory representation that compactly encodes spatial layouts and semantic contexts, effectively removing redundancies in raw episodic observations. 
A hierarchical chunking structure further organizes similar topo-polar trajectories into unified summaries, enabling coarse-to-fine retrieval. 
During navigation, candidate frontiers generate multiple trajectory hypotheses that query TrajRAG for similar past trajectories, guiding large-model reasoning for waypoint selection. 
New experiences are continually consolidated into TrajRAG, enabling the accumulation of lifelong navigation experience.
Experiments on MP3D, HM3D-v1, and HM3D-v2 show that TrajRAG effectively retrieves relevant geometric–semantic experiences and improves zero-shot ObjectNav performance.
\end{abstract}
    
\section{Introduction}
\label{sec:intro}



\begin{figure}[t]
\begin{centering}
\includegraphics[width=0.98\columnwidth]{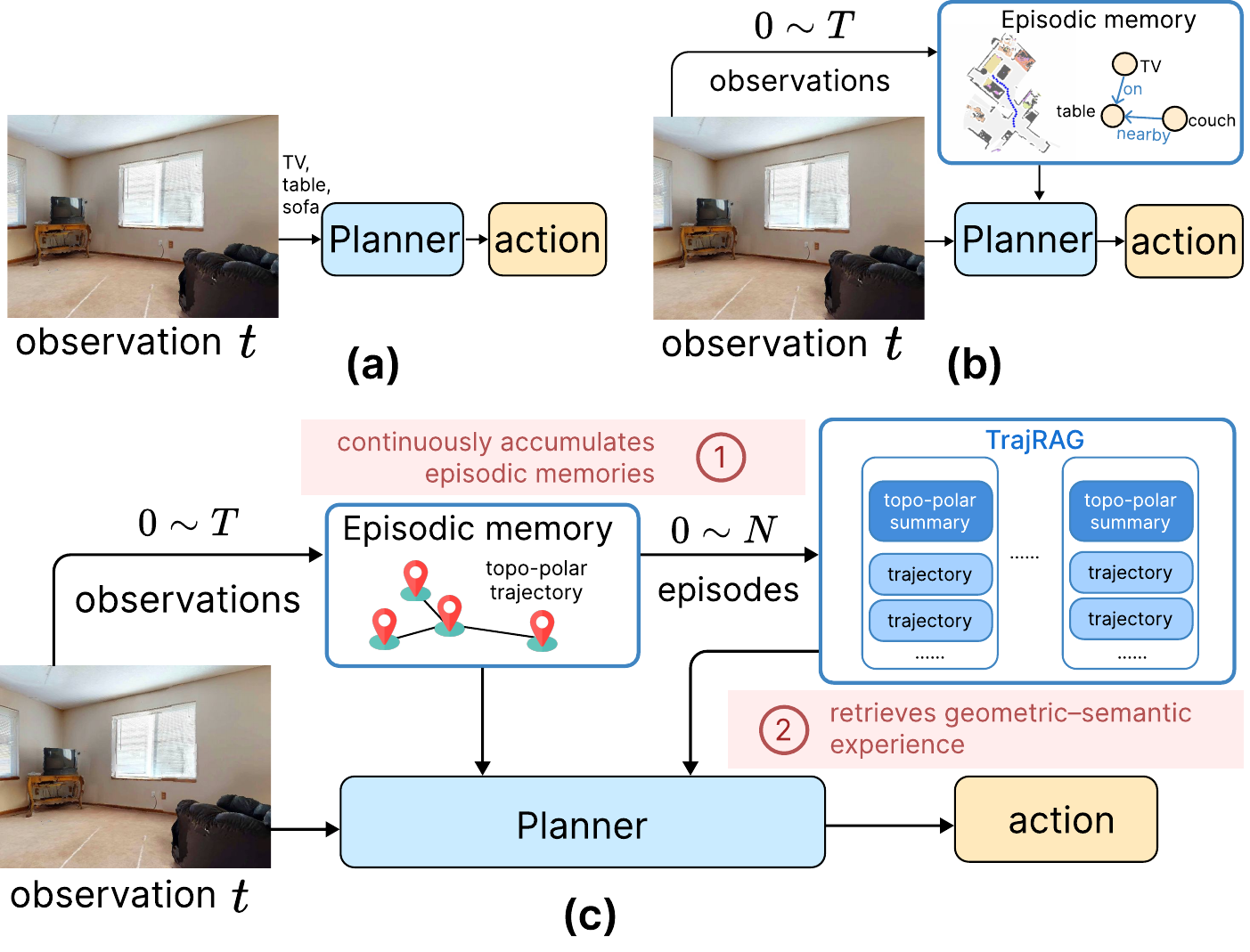}
\end{centering}
\vspace{-4pt}
\caption{\label{fig:into}
Comparisons with LLM/VLM-based Zero-shot ObjectNav methods.
(a) \textbf{Single-step context}. the planner (LLM/VLM) receives raw textual observations from the single timestep.
(b) \textbf{Episodic context}. episodic memory is structured into prompts for reasoning but discarded after each episode.
(c) \textbf{Episode + experience context}. our TrajRAG serves as long-term memory that continuously accumulates episodic memory and retrieves geometric–semantic experience for planning.
}
\vspace{-20pt}
\end{figure}

Object Goal Navigation (ObjectNav) requires an embodied agent to locate a user-specified object category in an unseen environment, based only on egocentric RGB-D observations. 
To navigate efficiently without exhaustive exploration, the agent needs to acquire semantic–geometric priors that encode how objects are typically arranged within spatial layouts. 
Prior works train models in seen environments to learn such priors, including object co-occurrence relations~\cite{hoz++PAMI,ECCV_relation_graph,DAT}, RL-based navigation policies~\cite{Goat-bench,RIM_Shizhe,ProcTHOR}, object-location prediction functions~\cite{PEANUT,PONI,L2M, wang2024lookahead, zsx_SGM}, and feature and layout adaptations for unseen environments~\cite{zsx_ECCV22,L-sTDE}. 
Recently, large-scale pretrained models (LLMs and VLMs) have demonstrated strong zero-shot common-sense reasoning abilities, inspiring methods that leverage them as general priors to perform ObjectNav in a zero-shot manner.


Existing LLM/VLM-based methods for zero-shot ObjectNav as illustrated in Fig.~\ref{fig:into}. 
Some methods (Fig.~\ref{fig:into}a)~\cite{Navgpt_aaai24,cai2024bridging,Cows_cvpr23} directly query large models with observations at each timestep for action planning, but the absence of episodic memory often results in redundant exploration and revisiting previously seen areas.
Other methods (Fig.~\ref{fig:into}b) introduce structured episodic memories, e.g., similarity maps~\cite{VLFM,FBN_iccv25,beliefmapnav,GAMap_nips24}, scene graphs~\cite{SG-Nav-nips24,unigoal_cvpr25,FBN_iccv25}, landmark graphs~\cite{VoroNav_icml24}, or 3D-language feature fields~\cite{wang2025g3d}, providing richer semantic cues for reasoning.
However, while these episodic memories are scene-specific, the large models used for decision-making rely on scene-agnostic knowledge learned from web-scale text rather than 3D spatial experience.  Furthermore, such episodic memories are typically discarded after each episode, limiting their scalability toward lifelong navigation.
In contrast, human navigation relies on both short-term and long-term memory~\cite{niv2019learning,elston2025context,whittington2022build}: the former encodes immediate environmental details, while the latter retrieves related experiences to support decision-making. 
Moreover, short-term memories are gradually consolidated into long-term memory~\cite{lisman2009prediction,momennejad2020learning,tang2023geometric}, enabling continual learning~\cite{rueckemann2021grid}.
To emulate this capability, an embodied agent requires a systematic internal representation~\cite{jiang2026self}, to continuously accumulate past experiences for lifelong navigation.


Motivated by this, we propose Trajectory RAG (TrajRAG), which aims to build a ``long-term'' memory of navigation experience (Fig.~\ref{fig:into}c) that (1) continuously accumulates episodic memories and (2) retrieves geometric–semantic experience to enhance the reasoning of large models.
For the first objective, raw episodic memories (trajectories of RGB-D observations) are redundant both within trajectories (due to revisits or local loops) and across trajectories (from spatial overlap among runs in the same scene).
To achieve memory compactness, we propose the topological-polar (topo-polar) trajectory to organize these raw trajectories.
Specifically, we first build a semantic map based on RGB-D observations, then skeletonize the navigable region to extract topological nodes. 
Around each node, space is discretized into polar sectors (30° per sector), each sector records the observed semantic labels. 
In the incremental construction of TrajRAG, the topo-polar trajectory offers two benefits: 
(1) The topological skeleton provides a structured representation of raw trajectories, enabling self-checking through the detecting and pruning of redundant segments, ensuring stored trajectories are efficient; and
(2) The polar-sector node captures relative geometric–semantic layouts as a distinctive fingerprint for matching new trajectories with existing ones in TrajRAG, determining whether the new one is redundant. This guarantees that the content of TrajRAG remains compact and informative.
Compared with other representations, the topo-polar trajectory achieves more accurate spatial matching than scene graphs, while being more flexible and requiring less computation cost than map-based or point-cloud–based matching.


For the second objective, TrajRAG adopts a hierarchical chunking architecture to enable efficient retrieval, in which each chunk corresponds to a topo-polar trajectory. 
Chunks with similar topological structures are grouped and merged into a unified topo-polar summary, serving as a descriptor for the group’s geometric–semantic layout.
During retrieval, a coarse-to-fine strategy is employed. 
Coarse matching between the query and topo-polar summaries first identifies relevant groups with similar layout, followed by fine-grained matching within these groups to locate the similar trajectory chunks.
As the number of trajectory chunks far exceeds that of summaries, a trajectory encoder is proposed to accelerate fine-grained retrieval. 
It embeds topo-polar trajectories, ensuring that trajectories from the same group with the same navigation goal lie close in the embedding space, while unrelated ones remain distant.

In this paper, we propose TrajRAG, a retrieval-augmented generation framework that retrieves geometric–semantic experiences for zero-shot object navigation.
During TrajRAG inference in navigation, the agent incrementally constructs a semantic map and transforms it into a topo-polar trajectory. At each timestep, candidate exploration points are selected as the centers of individual frontiers. Based on these candidates and the existing topo-polar trajectory, a set of candidate trajectories is generated. These candidate trajectories then perform coarse-to-fine retrieval over the TrajRAG, retrieving trajectory chunks that contain relevant geometric–semantic experiences. The retrieved experiences are provided to assist the policy model (e.g., LLMs) in selecting the next waypoint for exploration.
After completing an episode, the entire trajectory is incorporated into the TrajRAG, enabling lifelong experience accumulation.
We evaluate TrajRAG on zero-shot ObjectNav benchmarks across MP3D, HM3D-v1, and HM3D-v2. 
Experimental results demonstrate that integrating TrajRAG improves navigation accuracy and efficiency.
Furthermore, we validate the effectiveness of TrajRAG, demonstrating that it can effectively retrieve relevant geometric–semantic experiences during the decision-making process of large models. This capability bridges scene-agnostic commonsense reasoning and scene-specific experience, thereby enhancing generalization in zero-shot ObjectNav.

\section{Related Works}
\subsection{Zero-shot Object Navigation}


In zero-shot object navigation, with the rise of pre-trained models, recent methods utilize knowledge from pre-trained models to assist navigation decision-making and can be broadly categorized into two types: those leveraging Large Language Models (LLMs) and Vision-Language Models (VLMs)~\cite{Cows_cvpr23,ZSON,VLFM,ESC_ICML23}.
LLM-based methods primarily leverage the reasoning capabilities of LLMs to assist navigation. They extract and represent environmental information as natural-language structures, enabling the LLM to infer potential target locations and select the next exploration waypoint~\cite{VoroNav_icml24,Navgpt_aaai24,l3mvn,unigoal_cvpr25}. 
For instance, VoroNav~\cite{VoroNav_icml24} represents traversable paths as candidate trajectories and formulates the objects and scenes along these trajectories into natural-language descriptions, allowing the LLM to reason about and select the optimal path. 
Beyond using LLMs to infer target locations or assisting path selection, CogNav~\cite{Cao_2025_ICCV} elevates the LLM to determine the current navigation phase of the agent, thereby deciding which exploration policy the agent should execute.
VLM-based methods~\cite{VLFM,GAMap_nips24,beliefmapnav} extracts frontier points on a semantic map; then the VLM converts the similarity between frontier observations and goal text into frontier values, thereby guiding path planning.
Recent advancements empower VLMs with 3D understanding; e.g., Dynam3D~\cite{wangdynam3d} utilizes dynamic 3D tokens to provide long-term memory within a single environment.


Although prior methods use pretrained models, their knowledge is not grounded in real scene experience and cannot accumulate transferable memory. We propose TrajRAG, which retrieves geometric–semantic experiences from past trajectories and enables long-term memory through the continual accumulation of episodic experiences.

\subsection{Retrieval-Augmented Generation}

Retrieval-Augmented Generation (RAG)~\cite{lewis2020retrieval} was initially introduced to enhance large language models (LLMs) by retrieving relevant document fragments, thereby injecting domain-specific knowledge to improve factuality and relevance. 
Traditional RAG frameworks embed user queries and document chunks into a shared vector space, retrieving the top-k semantically similar passages to expand the model’s context window~\cite{gao2023precise,chan2024rq}. 
Recent RAG works introduced iterative retrieval and knowledge-graph-based methods. Models such as GraphRAG~\cite{edge2024local} and LightRAG~\cite{guo2024lightrag} extract entities and relations to construct graph structures, enabling more holistic and globally-aware retrieval.

Recently, researchers have begun adapting RAG to embodied intelligence~\cite{xie2024embodied, booker2024embodiedrag, wang2025navrag}. 
EmbodiedRAG~\cite{booker2024embodiedrag} and Embodied-RAG~\cite{xie2024embodied} build non-parametric, structured memory for the current scene in an online fashion, organizing scene information via RAG to enhance retrieval efficiency and accuracy over both the present environment and previously collected historical scenes.
NavRAG~\cite{wang2025navrag} proposes a retrieval-augmented framework that automatically generates user-oriented navigation instructions from structured simulator data without manual annotation, producing large-scale, high-quality instruction–trajectory pairs for Vision--Language--Navigation training, assuming that comprehensive scene information is already available.

Existing embodied-RAG approaches retrieve within the current scene and cannot transfer knowledge or leverage historical experience. TrajRAG instead continually indexes knowledge along novel trajectories and uses a coarse-to-fine retrieval pipeline to inject relevant past experience into the LLM, enabling cross-scene informed decisions.
\section{Method}



\subsection{TrajRAG Definition}

Raw episodic memories, represented as trajectories of RGB-D observations $\mathcal{T} = \{(I_t, D_t, \mathbf{p}_t)\}_{t=1}^T$ (where $I_t$, $D_t$, and $\mathbf{p}_t$ denote RGB, depth, and agent pose, respectively), are inherently redundant due to local revisits, small inter-frame variations, and cross-episode overlaps.  
To achieve compact yet semantically meaningful memory, we propose the topological–polar (topo-polar) trajectory, which transforms continuous raw observations into a structured representation.

\textbf{Topo-polar trajectory}.
Given the raw observations $\mathcal{T} = \{(I_t, D_t, \mathbf{p}_t)\}_{t=1}^T$, the map module first constructs an open-vocabulary semantic map $\mathcal{M}_t$ (see Sec.~\ref{sec:navigation}).
Based on the explored regions in the semantic map $m_t^{free} \in \mathcal{M}_t$, a morphological thinning operation $\mathcal{S}(\cdot)$ is applied to obtain a one-pixel-wide skeleton, formally $\mathcal{G}_{\text{skel}} = \mathcal{S}(m_t^{free})$.

Then, candidate topological nodes are defined by selecting pixels whose 8-neighborhood contains at least three connected components:
\begin{equation}
\mathcal{V}_{\text{cand}} = \{ v \in \mathcal{G}_{\text{skel}} \mid |\mathcal{N}_8(v)| \ge 3 \}
\end{equation}
Since candidate nodes $\mathcal{V}_{\text{cand}}$ may exhibit spatial redundancy from local skeleton noise or nearby junctions, we apply non-maximum distance suppression, merging nodes within a threshold distance to form the final node set $\mathcal{V} = \{v_k\}_{k=1}^{N_v}$ that defines the discrete topology of the explored region.

For each node $v_k$, we adopt a polar-coordinate representation to characterize its surrounding geometric-semantic context. 
Since different trajectories arise from varying agent initial poses, the absolute positions of observed objects are inconsistent across runs. In contrast, the polar representation captures relative spatial relationships, making it more suitable for this setting. Moreover, the division of space into multiple angular sectors allows for a clear distinction of node-specific characteristics, which facilitates subsequent node matching.
Specifically, for each node $v_k$, we cast polar sampling rays centered at its position in $\mathcal{M}_t$.  
Each ray $\mathcal{R}(\theta)$, with direction $\theta \in [0, 2\pi)$ and step $\Delta\theta = 30^{\circ}$, searches for the first non-free pixel (i.e., an object, obstacle, or unknown region) within range $R$. 
The sampling function is defined as:
\begin{equation}
\phi_k(\theta) \!\!=\!\!
\begin{cases}
c, & \!\! \small{\text{if hits object } c} \\
\text{obstacle}, & \!\! \small{\text{if hits obstacle}} \\
\text{unknown}, & \!\! \small{\text{if hits unknown region}} \\
\text{free}, & \!\! \small{\text{if no hit within } R}
\end{cases}
\end{equation}
Before sampling, each semantic channel is morphological dilation to prioritize semantic pixels are prioritized over obstacles, preventing incorrect associations caused by thin wall boundaries.  
The sector vector around $v_k$ is:
\begin{equation}
\mathbf{s}_k = [\phi_k(\theta_1), \phi_k(\theta_2), \ldots, \phi_k(\theta_{12})]
\label{eq:polar node}
\end{equation}
Sector indices increase counterclockwise relative to the agent’s left-facing direction.  
This vector $\mathbf{s}_k$ encodes local semantic–geometric structure, and episode-specific world coordinate $\mathbf{p}_k$ of each node $v_k$ is recorded for alignment.

For each segment $\tau = (I_t, D_t, \mathbf{p}_t)$ in the observations $\mathcal{T} = \{(I_t, D_t, \mathbf{p}_t)\}_{t=1}^T$, we normalize it by assigning it to its nearest node according to the agent’s pose measured by Euclidean distance:
\begin{equation}
v_t^* = \arg\min_{v_k \in \mathcal{V}} \|\mathbf{p}_t - \mathbf{p}_k\|_2
\end{equation}
Consecutive identical nodes are first merged to eliminate short-term revisits within the same topological region. 
To further remove long-term cycles, we prune loops (denoted as $f_{\text{PL}}$) by retaining only the last occurrence of each node in temporal order, i.e., when a node reappears, all nodes between its previous and latest occurrence are discarded. 
This ensures that the stored trajectories are loop-free and efficient, avoiding redundancy within each individual trajectory.
The resulting sequence defines a topo-polar trajectory $\mathcal{T}_{\text{tp}}$ for the raw observations $\mathcal{T}$:
\begin{equation}
\mathcal{T}_{\text{tp}} = (\mathcal{V},\mathcal{E}),\quad \mathcal{V}=f_{\text{PL}}(\{v_t^*\})
\label{eq:topo-polar trajectory}
\end{equation}
where $\mathcal{E}$ denotes directed edges between consecutive nodes.

\textbf{TrajRAG structure.}
TrajRAG adopts a hierarchical chunking architecture to organize the topo-polar trajectories, consisting of the following key components:

1) Chunk. Each chunk $\chi_i$ corresponds to a topo-polar trajectory, which includes a topological–polar trajectory $\mathcal{T}_{\text{tp}}^i$, a natural language description $L(\mathcal{T}_{\text{tp}}^i)$, and the corresponding trajectory embedding $\mathbf{z}_i = f_{\text{E}}(\mathcal{T}_{\text{tp}}^i)$.

2) Coarse index. Chunks with high geometric–semantic similarity are grouped and merged into a unified topo-polar summary. The similarity between two chunks is measured by the geometric–semantic consistency between their topo-polar trajectories (see Sec.~\ref{sec:RAGconstruct} for details). The topo-polar summary is defined as $\mathcal{G}_{\text{sum}} = (\mathcal{V}_{\text{uni}}, \mathcal{E}_{\text{mrg}})$, 
where $\mathcal{V}_{\text{uni}}$ and $\mathcal{E}_{\text{mrg}}$ denote the fused nodes and edges after removing duplicates.
The summary graph $\mathcal{G}_{\text{sum}}$ captures the group’s overall geometric–semantic layout and serves as a \textbf{coarse index} during retrieval. Given a query, TrajRAG first retrieves a set of relevant summaries before performing fine-grained trajectory search. This process narrows down the trajectory search to relevant groups, enhancing the relevance of the retrieved historical trajectory knowledge to the current scene.

3) Fine index.
Since the number of chunks is much larger than the number of topo-polar summaries, each chunk is encoded as a hidden vector embedding to enable efficient retrieval during real-time navigation.
Each topo-polar node $v_k$ is represented by its sector vector $\mathbf{s}_k$.
We employ an encoder-only transformer (e.g., DistilBERT) to obtain node embeddings, as its bidirectional attention and position-insensitive properties make it well-suited for encoding polar representations: $\mathbf{h}_k = \mathcal{E}_{\text{node}}(\mathbf{s}_k)$.
Given the ordered sequence of node embeddings $[\mathbf{h}_1, \mathbf{h}_2, \dots, \mathbf{h}_L]$, we encode their sequential dependencies using a decoder-only transformer (e.g., DistilGPT2) denoted as $\mathcal{D}_{\text{traj}}$.
The order-sensitive $\mathcal{D}_{\text{traj}}$ captures temporal correlations along the traversal order.
We then concatenate the final token representation with the goal semantic embedding $\mathbf{h}_{\text{g}}$ to form the trajectory embedding:
\begin{equation}
\mathbf{z} = f_{\text{E}}(\mathcal{T}_{\text{tp}}) \!\!= \!\! \small{\mathbf{h}_L'\oplus\mathbf{h}_{\text{g}}, \quad
[\mathbf{h}_1', ..., \mathbf{h}_L']\!\!=\!\!\mathcal{D}_{\text{traj}}\big([\mathbf{h}_1,..., \mathbf{h}_L]\big)}
\end{equation}
where $\oplus$ denotes concatenation.
We optimize the trajectory encoder $f_{\text{E}}$ (including both the encoder-only and decoder-only transformers) using a contrastive learning objective:
\begin{equation}
\mathcal{L}_{\text{contrast}} = - \log
\frac{\exp(\text{sim}(\mathbf{z}_i, \mathbf{z}_j^+)/\tau)}
{\sum_{k}\exp(\text{sim}(\mathbf{z}_i, \mathbf{z}_k)/\tau)}
\end{equation}
where $\text{sim}(\mathbf{z}_i,\mathbf{z}_j)=\mathbf{z}_i^\top\mathbf{z}_j/(\|\mathbf{z}_i\|\|\mathbf{z}_j\|)$ denotes cosine similarity and $\tau$ is a temperature parameter.  
Positive pairs $(\mathbf{z}_i,\mathbf{z}_j^+)$ are sampled from trajectories within the same topological group or sharing the same navigation goal, while negatives are randomly drawn from different groups.
Note that during training, the encoder-only transformer is initialized with pretrained weights and kept frozen, i.e., its parameters are not updated.

\begin{figure*}[t]
\begin{centering}
\includegraphics[width=0.99\textwidth]{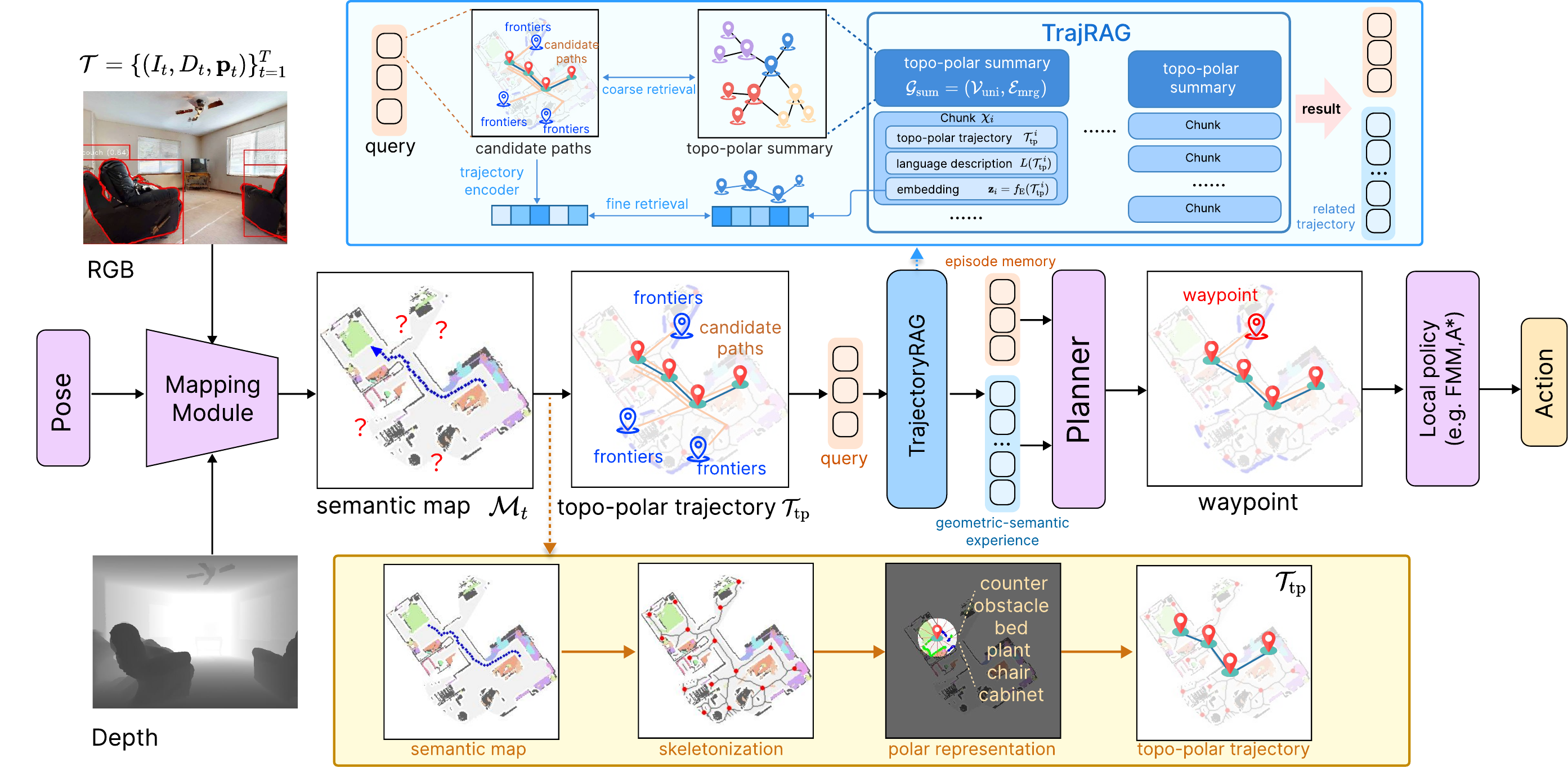}
\par\end{centering}
\caption{
\textbf{Navigation Framework of TrajRAG.}
The agent incrementally maintains a semantic map during navigation. Based on this map, we convert the explored area into a topo-polar trajectory. Candidate trajectories are then generated according to the potential frontiers. For each candidate, TrajRAG retrieves relevant experiences to help the planner estimate which trajectory can reach the goal more efficiently. A coarse-to-fine retrieval strategy is adopted to accelerate retrieval. 
}
\label{fig:framework}
\vspace{-10pt}
\end{figure*}

\subsection{Incremental Construction\label{sec:RAGconstruct}}

During navigation, TrajRAG is incrementally constructed.  
Given a new episode $\mathcal{T} = \{(I_t, D_t, \mathbf{p}_t)\}$, we first derive its topo-polar trajectory $\mathcal{T}_{\text{tp}}$ as defined in Eq.~\ref{eq:topo-polar trajectory}.
The resulting trajectory is then matched against existing topo-polar summaries for potential integration.

\textbf{Semantic matching}.
We compute the semantic similarity matrix $\mathbf{S} \in \mathbb{R}^{N_{\text{new}} \times N_{\text{sum}}}$ between the nodes of $\mathcal{T}_{\text{tp}}$ and those in each summary graph:
\begin{equation}
S_{ij} = \max_{\Delta\theta} \text{sim}\big(\text{Rot}(\mathbf{s}_i, \Delta\theta), \mathbf{s}_j\big)
\label{eq:semantic matching}
\end{equation}
where $\text{Rot}(\mathbf{s}_i, \Delta\theta)$ cyclically rotates the 12-dimensional sector vector to compensate for heading misalignment, which ensures that nodes with similar surrounding semantics are matched even when the agent’s facing direction differs across trajectories. 
After computing $\mathbf{S}$, we adopt a bidirectional $K$-nearest-neighbor (mutual KNN) strategy to select consistent correspondences, i.e., a pair $(v_i, v_j)$ is retained only if $v_j$ ranks among the top-$K$ nearest neighbors of $v_i$ and vice versa.  
This symmetric constraint filters out spurious matches while preserving high-confidence node pairs for subsequent geometric matching.

\textbf{Geometric matching}.
Using the world coordinates $\{\mathbf{p}_k\}$ stored in matched nodes, we apply RANSAC~\cite{fischler1981random} estimation to find a robust geometric transformation $\mathbf{T} \in SE(2)$ aligning the new trajectory to the summary:
\begin{equation}
\mathbf{T} = \arg\min_{\mathbf{T}} \sum_{(v_i, v_j) \in \mathcal{C}} \rho\big(\|\mathbf{T}\mathbf{p}_i - \mathbf{p}_j\|\big)
\label{geno matching}
\end{equation}
where $\mathcal{C}$ is the set of matched node pairs and $\rho(\cdot)$ is a robust penalty.  
We set the geometric similarity score to the inlier ratio $\vert\mathcal{C}_{\text{in}}\vert/\vert\mathcal{C}\vert$ returned by RANSAC, with $\mathcal{C}{\text{in}}$ defined as the set of consistent pairs under the estimated transformation.
If a valid $\mathbf{T}$ is found, $\mathcal{T}_{\text{tp}}$ is merged into the corresponding summary group, extending $\mathcal{V}_{\text{unique}}$ and $\mathcal{E}_{\text{merged}}$ with newly discovered nodes and edges.

\textbf{Redundancy checking and TrajRAG update}.
After semantic and geometric matching, we further compare the new trajectory against existing members within each group to prevent redundant storage. 
Specifically, if two trajectories share the same navigation goal and their node sequences exhibit a strict containment relationship (i.e., one trajectory’s node list is a subsequence of the other with identical order), the shorter one is regarded as redundant and discarded. 
Otherwise, the new trajectory enriches the group by contributing additional geometric-semantic context.  

Finally, the group’s summary graph $\mathcal{G}_{\text{sum}}$ is updated by integrating the new trajectory.
Given the estimated transformation $\mathbf{T} \in SE(2)$, the trajectory’s nodes and edges are first transformed into the global frame. Newly discovered nodes and edges absent from the existing summary are then identified and added, allowing $\mathcal{G}{\text{sum}}$ to expand incrementally as navigation progresses.
Additionally, if a new topo-polar trajectory cannot be matched with any existing group in terms of geometric and semantic similarity, a new group is created, and this trajectory is registered as the group’s initial topo-polar summary and added to TrajRAG.

We collected navigation trajectories on the training splits of HM3Dv1~\cite{ramakrishnan2021habitat} and MP3D~\cite{MP3D_data} to build TrajRAG.
From the HM3Dv1 ObjectNav Train dataset (3.9M episodes, 6 categories, 80 scenes), we uniformly sample over 200k trajectories, ensuring exhaustive coverage of all floors and categories per scene. 
Similarly, from the MP3D ObjectNav Train dataset (2.6M episodes, 21 categories, 56 scenes), we gather more than 150k trajectories, again achieving coverage of all floor layouts and object categories in every scene.

\subsection{Navigation with TrajRAG\label{sec:navigation}}

In the zero-shot ObjectNav task, an agent is initialized in an unseen environment and must locate an object from a specified semantic category (e.g., chair) through active exploration, using real-time RGB-D observations and sensor pose. Success requires issuing \texttt{stop} within a distance threshold (e.g., 1 m) of the target and within the step budget (e.g., 500 steps).

As shown in Fig.~\ref{fig:framework}, navigation with TrajRAG comprises constructing the current-scene trajectory and retrieving and exploiting trajectory knowledge from the knowledge base.
At timestep \( t \) of an episode, we utilize a \textbf{semantic map module} to incrementally construct a semantic map based on RGB-D observations and agent poses. 
GroundingDINO~\cite{GroundingDINO} detects object categories, and MobileSAM~\cite{mobileSAM} provides more precise segmentation masks.
The semantic map is formulated as \( \mathcal{M}_t \in \mathbb{R}^{(2 + N_o) \times H \times W} \), 
where \( N_o \) denotes the number of object categories, which can be arbitrarily specified. 
The remaining two channels correspond to obstacles and explored areas.

Based on the semantic map, a topo-polar trajectory is constructed following Eq.~\ref{eq:topo-polar trajectory}. 
From the obstacles and explored areas of the semantic map, frontiers are extracted, defined as the boundaries between explored and unexplored areas. 
Each frontier is represented by its geometric centroid and filtered to remove short and noisy segments. 
Similar to the node in topo-polar trajectory, each frontier is further represented by a sector vector as defined in Eq.~\ref{eq:polar node}. 
A breadth-first search (BFS) is performed over the current topo-polar trajectory to compute paths leading to all valid frontiers. 
This process yields candidate paths \( \{\Pi_i\}_{i=0}^{N} \).

To provide the navigation planner with experience on what typically follows a candidate path (i.e., which path may lead closer to the goal), we perform hierarchical retrieval of these candidate trajectories within TrajRAG to locate the most relevant experiences.
Specifically, a coarse retrieval stage first identifies top-ranked similar groups by comparing stored topo-polar summaries with the current candidate topo-polar trajectory, using both semantic and geometric matching results (as described in Eq.~\ref{eq:semantic matching} and Eq.~\ref{geno matching}).

Within the selected groups, a fine-grained retrieval stage encodes the current topo-polar trajectory with the trajectory encoder $f_{E}$, and retrieves the top-$K$ most similar trajectories.
For each retrieved trajectory $\mathcal{T}_j$, we utilize its corresponding language description $\mathcal{L}_j$ as the experience. 
The current candidate paths $\{\Pi_i\}$ and retrieved experience $\{\mathcal{L}_j\}$ are jointly provided to a navigation planner (e.g., LLM-based model) to select the optimal navigation path $\Pi^*$.
Whenever the agent reaches a new topo-polar node, the planner performs an inference.
The frontier corresponding to $\Pi^*$ is then set as the next waypoint for the local policy.
Our local policy computes the shortest path from the current position to the waypoint using the A* algorithm, and then determines the next action for the agent by following this trajectory through its discrete action space.

\section{Experiments}

\subsection{Experimental Setup}

\textbf{Datasets and Metrics}.
For \textbf{zero-shot ObjectNav}, we evaluate our method on the HM3Dv1~\cite{ramakrishnan2021habitat}, HM3Dv2~\cite{HM3D_data} , and MP3D~\cite{MP3D_data} within the Habitat simulator~\cite{Habitat}. 
Following previous works~\cite{ApexNAV_RAL25,unigoal_cvpr25}, we report Success Rate (SR) and Success weighted by normalized inverse Path Length (SPL) as the evaluation metrics.

\textbf{Implementation Details}.
During navigation, we employ GroundingDINO~\cite{GroundingDINO} and MobileSAM~\cite{mobileSAM} to detect and segment images and generate a semantic map, while Qwen3-32B~\cite{qwen3technicalreport} is used as the planner to select the path. 
In the free map, topological keypoints are deduplicated with a distance threshold $d_{\min}$ of 0.5 meters. For generating textual descriptions of each sector in the semantic map, a sampling radius of 1.5 meters is employed.
To ensure fair evaluation and prevent test-set leakage, TrajRAG is pre-built solely on training data and kept frozen during testing, although it natively supports dynamic test-time updates.

\subsection{Evaluation Results}

\begin{table}
\setlength{\tabcolsep}{10pt} \renewcommand{\arraystretch}{1.1}
\caption{\label{tab:ablation-node-representations}Ablation on node representations in HM3Dv1. TNA: textual neighbor aggregation; TPS-G: topo-polar sector geometry; TPS-S: topo-polar sector semantics.}
\vspace{-5pt}
\centering{}%
\begin{tabular}{ccc|cc}
\hline 
{\footnotesize TNA} & {\footnotesize TPS-G} & {\footnotesize TPS-S} & {\footnotesize SR(\%)} & {\footnotesize SPL(\%)}\tabularnewline
\hline 
{\footnotesize$\checkmark$} &  &  & {\footnotesize 53.9} & {\footnotesize 25.7}\tabularnewline
 & {\footnotesize$\checkmark$} &  & {\footnotesize 48.1} & {\footnotesize 22.3}\tabularnewline
 &  & {\footnotesize$\checkmark$} & {\footnotesize 57.3} & {\footnotesize 30.6}\tabularnewline
\hline 
 & {\footnotesize$\checkmark$} & {\footnotesize$\checkmark$} & {\footnotesize \textbf{61.7}} & {\footnotesize \textbf{33.2}}\tabularnewline
\hline 
\end{tabular}
\vspace{-10pt}
\end{table}

\textbf{Ablation on Node Representations}.
Tab.~\ref{tab:ablation-node-representations} presents the ablation on node representations. 
Our full model, which integrates both Topo-Polar Sector geometry and semantics (TPS-G \& TPS-S), achieves the best performance, demonstrating the complementarity of geometric and semantic cues.
In contrast, the Textual Neighbor Aggregation (TNA) baseline (Line 1), which aggregates textual descriptions without spatial order for trajectory description and node embedding, performs worse.
Furthermore, using only Topo-Polar Sector-Geometry (TPS-G) yields the weakest performance, as geometry alone lacks semantic cues.
Employing only Topo-Polar Sector-Semantics (TPS-S) surpasses both TNA and TPS-G, demonstrating the importance of structured semantic description and encoding.
In summary, our topo-polar format provides a clear advantage by preserving spatial relationships and structuring semantic information.

\begin{table}
\setlength{\tabcolsep}{12pt} \renewcommand{\arraystretch}{1.1}
\caption{\label{tab:ablation-retrieve-strategy}Ablation study on retrieval strategy in HM3Dv1. TE: text embedding; SE: our sequence embedding.}
\vspace{-5pt}
\centering{}%
\begin{tabular}{cc|cc}
\hline 
{\footnotesize Coarse} & {\footnotesize Fine} & {\footnotesize SR(\%)} & {\footnotesize SPL(\%)}\tabularnewline
\hline 
 & {\footnotesize SE} & {\footnotesize 54.3} & {\footnotesize 25.6}\tabularnewline
{\footnotesize$\checkmark$} & {\footnotesize TE} & {\footnotesize 57.8} & {\footnotesize 29.7}\tabularnewline
\hline 
{\footnotesize$\checkmark$} & {\footnotesize SE} & {\footnotesize \textbf{61.7}} & {\footnotesize \textbf{33.2}}\tabularnewline
\hline 
\end{tabular}
\vspace{-10pt}
\end{table}

\textbf{Ablation Study on the Retrieval Strategy}.
Tab.~\ref{tab:ablation-retrieve-strategy} evaluates the impact of each component in our retrieval strategy. TrajRAG employs a coarse-to-fine representation and retrieval pipeline. The coarse stage retrieves scenes based on topo-polar geometric-semantic relationships, while the fine stage retrieves specific trajectories using embeddings.
In the first row, we remove coarse-level matching between the current scene and database summaries, retrieving trajectories directly with sequence embeddings. This leads to a clear performance drop, confirming that pre-retrieval based on scene layout is essential for filtering irrelevant contexts effectively.
The second row retains scene-level matching but replaces our trained trajectory encoder with a pre-trained text model to encode structured trajectory descriptions. This also degrades results, underscoring the limitation of generic text embeddings compared with our dedicated sequence encoder in capturing trajectory semantics.
The best performance is achieved by integrating both retrieval stages, which validates the design of TrajRAG’s hierarchical retrieval system.

\begin{figure}[t]
\begin{centering}
\includegraphics[width=0.98\columnwidth]{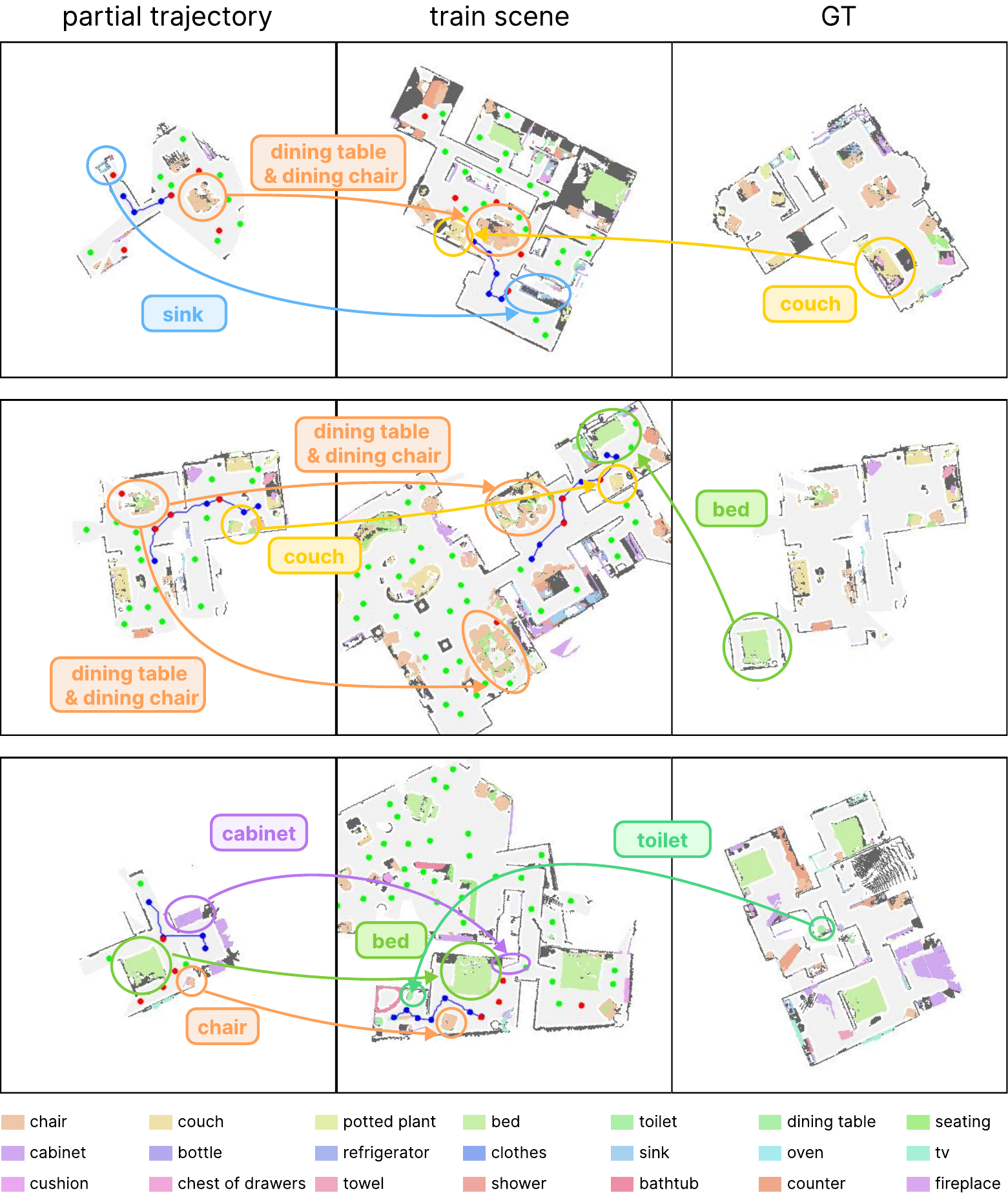}
\par\end{centering}
\vspace{-5pt}
\caption{\textbf{Retrieved Examples}. 
Given a partial trajectory, TrajRAG uses its accumulated keypoints (green) to retrieve a layout-consistent scene and the assistant goal-reaching trajectory (blue). For pre-retrieval, the top-5 matched keypoints within the visible map are highlighted in red.
}
\label{fig:retrieved-examples}
\vspace{-20pt}
\end{figure}

\begin{figure*}[t]
\begin{centering}
\includegraphics[width=0.98\textwidth]{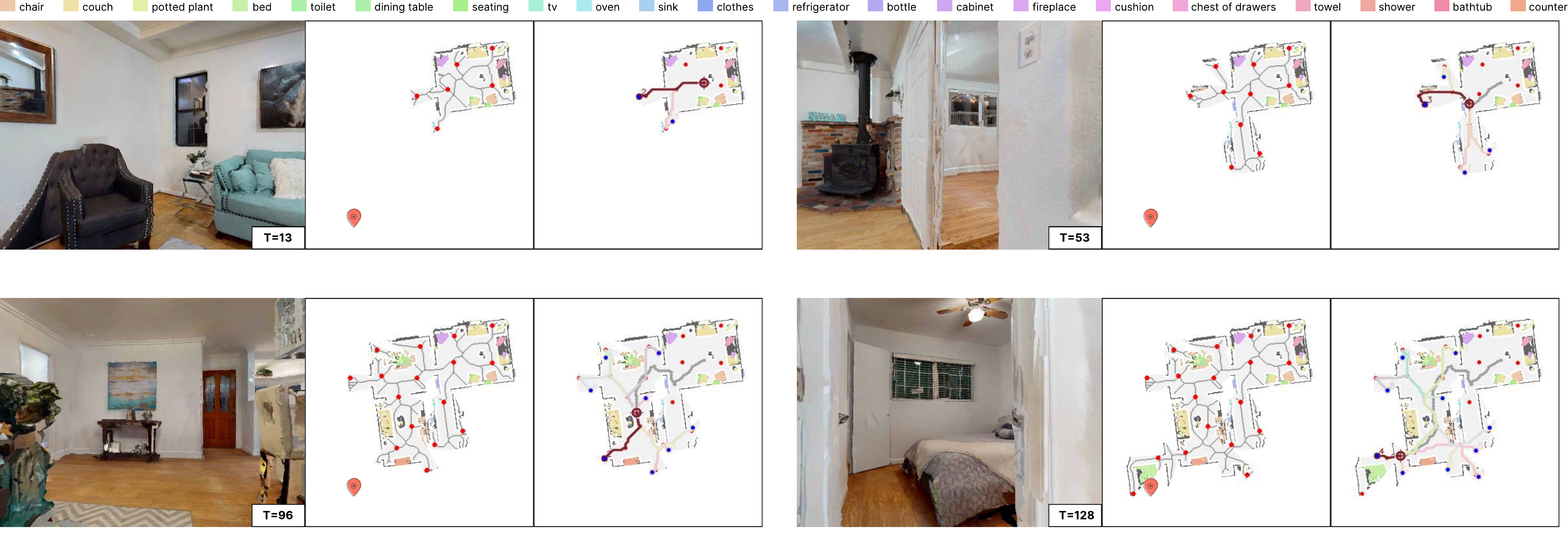}
\par\end{centering}
\caption{\textbf{Navigation with TrajRAG}. The left column shows the agent's ego-view RGB images. The middle column presents the skeletonization and detected keypoints (red) on the skeleton map, where the location icon indicates the ground-truth location of the target (``bed'') for visualization purposes only; the target location is unknown to the agent during navigation. The right column illustrates the agent's traversed trajectory from the start to the pre-navigation position (light gray), candidate trajectories from the current location to various frontiers (light-colored, with blue dots indicating frontiers), and the trajectory selected by the model (burgundy).
}
\label{fig:exp-navigation}
\vspace{-10pt}
\end{figure*}

\textbf{Qualitative Analysis}.
To better understand how our TrajRAG supports navigation decisions, we visualize the trajectory retrieval results in Fig.~\ref{fig:retrieved-examples}. The keypoints accumulated from the partial trajectory are used as the query for retrieval. We highlight the top-5 keypoints pre-retrieved based on scene layout (shown in red) as well as the retrieved trajectory leading toward the target object (shown in blue). As illustrated, the retrieved trajectory exhibits a strong latent spatial alignment with the test environment. Although it is retrieved from a different scene, its geometric structure and goal-directed progression are highly compatible with the query layout. This alignment indicates that TrajRAG captures scene-invariant spatial relations and retrieve trajectories that reflect how agents typically approach similar targets. Consequently, the retrieved trajectory provides meaningful guidance for downstream navigation, demonstrating the effectiveness of TrajRAG.

In addition, Fig.~\ref{fig:exp-navigation} shows that the agent can effectively choose the right waypoint based on the TrajRAG.



\begin{table}
\setlength{\tabcolsep}{1pt} \renewcommand{\arraystretch}{1.2}
\caption{\label{tab:RAG-comparison}Comparison with Other Types of RAG in HM3Dv1.}
\vspace{-5pt}
\centering{}%
\begin{tabular}{c|c|c|cc}
\hline 
\multicolumn{1}{c|}{{\footnotesize Method}} & \multicolumn{1}{c|}{{\footnotesize Retrieval}} & \multicolumn{1}{c|}{{\footnotesize Content}} & \multirow{1}{*}{{\footnotesize SR(\%)}} & \multirow{1}{*}{{\footnotesize SPL(\%)}}\tabularnewline
\hline 
{\footnotesize TrajTextRAG} & {\footnotesize text embedding} & {\footnotesize text description} & {\footnotesize 53.3} & {\footnotesize 25.6}\tabularnewline
{\footnotesize GraphRAG} & {\footnotesize graph embedding} & {\footnotesize textual scene graph} & {\footnotesize 55.2} & {\footnotesize 30.7}\tabularnewline
{\footnotesize TrajRAG(Ours)} & {\footnotesize hierarchical retrieval} & {\footnotesize topo-polar trajectory} & {\footnotesize \textbf{61.7}} & {\footnotesize \textbf{33.2}}\tabularnewline
\hline 
\end{tabular}
\vspace{-10pt}
\end{table}

\textbf{Comparison with other types of RAG.}
As shown in Tab.~\ref{tab:RAG-comparison}, we compare our method with two alternative RAG formulations. Our TrajRAG employs a hierarchical retrieval structure that accounts for scene layout, and retrieves topo-polar trajectories to assist the LLM in selecting suitable candidate paths.
TrajTextRAG encode historical trajectories as textual sequences and perform retrieval in a semantic embedding space. The method retrieves trajectory descriptions based on current candidate paths and uses them to guide the LLM. However, its performance is lower than ours, suggesting that the textual semantic space struggles to distinguish sequential relationships in trajectories and lacks scene-aware pre-retrieval, leading to suboptimal matching with the current scene.
GraphRAG construct scene graphs where objects serve as nodes and spatial relations (e.g., “next to”) as edges. During navigation, relevant scene graphs are retrieved to prompt the LLM for frontier scoring. This variant performs worse than TrajRAG. A plausible explanation for this gap is that the retrieved scene graphs often contain generic information and lack trajectory continuity, thus providing limited actionable knowledge for the LLM.

\begin{table}
\setlength{\tabcolsep}{4pt} \renewcommand{\arraystretch}{1}
\caption{\label{tab:cross-dataset}Cross-dataset evaluation results on HM3Dv1 and MP3D.}
\vspace{-5pt}
\centering{}%
\begin{tabular}{c|c|c|cc}
\hline 
 & {\footnotesize HM3Dv1 train} & {\footnotesize MP3D train} & {\footnotesize SR(\%)} & {\footnotesize SPL(\%)}\tabularnewline
\hline 
\multirow{3}{*}{{\footnotesize HM3Dv1 val}} & {\footnotesize$\checkmark$} &  & {\footnotesize 61.7} & {\footnotesize 33.2}\tabularnewline
 &  & {\footnotesize$\checkmark$} & {\footnotesize 59.8} & {\footnotesize 31.6}\tabularnewline
 & {\footnotesize$\checkmark$} & {\footnotesize$\checkmark$} & {\footnotesize \textbf{62.5}} & {\footnotesize \textbf{33.9}}\tabularnewline
\hline 
\multirow{3}{*}{{\footnotesize MP3D val}} & {\footnotesize$\checkmark$} &  & {\footnotesize 39.4} & {\footnotesize 16.2}\tabularnewline
 &  & {\footnotesize$\checkmark$} & {\footnotesize 41.7} & {\footnotesize 17.6}\tabularnewline
 & {\footnotesize$\checkmark$} & {\footnotesize$\checkmark$} & {\footnotesize \textbf{42.6}} & {\footnotesize \textbf{18.0}}\tabularnewline
\hline 
\end{tabular}
\vspace{-10pt}
\end{table}

\textbf{Cross-dataset Evaluation}.
Table~\ref{tab:cross-dataset} evaluates the cross-dataset generalization of our retrieval corpus. 
The results demonstrate that a retrieval corpus built from one dataset (e.g., HM3Dv1) generalizes effectively to guide navigation in another (e.g., MP3D), and vice versa, with only a minor performance drop compared to in-domain retrieval. This indicates that the semantic and topological representations in our retrieval corpus capture universal navigation cues that transfer across different environments. 
Furthermore, combining the retrieval corpora from both datasets consistently enhances performance on both validation sets, achieving the best results. 
This shows that our method not only transfers well but also benefits from a more diverse and comprehensive set of navigation experiences.

\subsection{Comparison with SOTA Methods}

\begin{table}
\setlength{\tabcolsep}{2pt} \renewcommand{\arraystretch}{1.1}
\caption{\label{tab:ObjectNav}Comparisons with the related works on MP3D, HM3Dv1 and HM3Dv2 datasets. ``OV'' denotes if the method supports open-vocabulary object goals.}
\centering{}%
\begin{tabular}{cccc|cc|cc}
\hline 
\multirow{2}{*}{{\footnotesize Method}} & \multirow{2}{*}{{\footnotesize OV}} & \multicolumn{2}{c|}{{\footnotesize MP3D}} & \multicolumn{2}{c|}{{\footnotesize HM3Dv1}} & \multicolumn{2}{c}{{\footnotesize HM3Dv2}}\tabularnewline
 &  & {\footnotesize SR(\%)} & {\footnotesize SPL(\%)} & {\footnotesize SR(\%)} & {\footnotesize SPL(\%)} & {\footnotesize SR(\%)} & {\footnotesize SPL(\%)}\tabularnewline
\hline 
{\footnotesize DD-PPO\cite{DD-PPO}} & {\footnotesize$\times$} & {\footnotesize 8.0} & {\footnotesize 1.8} & {\footnotesize 27.9} & {\footnotesize 14.2} & {\footnotesize -} & {\footnotesize -}\tabularnewline
{\footnotesize SemExp \cite{Chaplot_NIPS20}} & {\footnotesize$\times$} & {\footnotesize 36.0} & {\footnotesize 14.4} & {\footnotesize -} & {\footnotesize -} & {\footnotesize -} & {\footnotesize -}\tabularnewline
{\footnotesize SGM \cite{zsx_SGM}} & {\footnotesize$\times$} & {\footnotesize 37.7} & {\footnotesize 14.7} & {\footnotesize 60.2} & {\footnotesize 30.8} & {\footnotesize -} & {\footnotesize -}\tabularnewline
{\footnotesize T-Diff \cite{T-Diff}} & {\footnotesize$\times$} & {\footnotesize 39.6} & {\footnotesize 15.2} & {\footnotesize -} & {\footnotesize -} & {\footnotesize -} & {\footnotesize -}\tabularnewline
{\footnotesize GOAL \cite{GOAL_nips25}} & {\footnotesize$\times$} & {\footnotesize 41.7} & {\footnotesize 15.5} & {\footnotesize -} & {\footnotesize -} & {\footnotesize -} & {\footnotesize -}\tabularnewline
\hline 
{\footnotesize ZSON \cite{ZSON}} & {\footnotesize$\checkmark$} & {\footnotesize 15.3} & {\footnotesize 4.8} & {\footnotesize 25.5} & {\footnotesize 12.6} & {\footnotesize -} & {\footnotesize -}\tabularnewline
{\footnotesize PSL \cite{PSL_eccv24}} & {\footnotesize$\checkmark$} & - & - & {\footnotesize 42.4} & {\footnotesize 19.2} & {\footnotesize -} & {\footnotesize -}\tabularnewline
{\footnotesize ESC \cite{ESC_ICML23}} & {\footnotesize$\checkmark$} & {\footnotesize 28.7} & {\footnotesize 14.2} & {\footnotesize 39.2} & {\footnotesize 22.3} & {\footnotesize -} & {\footnotesize -}\tabularnewline
{\footnotesize VLFM \cite{VLFM}} & {\footnotesize$\checkmark$} & {\footnotesize 36.4} & {\footnotesize 17.5} & {\footnotesize 52.5} & {\footnotesize 30.4} & {\footnotesize 63.6} & {\footnotesize 32.5}\tabularnewline
{\footnotesize VoroNav \cite{VoroNav_icml24}} & {\footnotesize$\checkmark$} & {\footnotesize -} & {\footnotesize -} & {\footnotesize 42.0} & {\footnotesize 26.0} & {\footnotesize -} & {\footnotesize -}\tabularnewline
{\footnotesize InstructNav \cite{InstructNav}} & {\footnotesize$\checkmark$} & {\footnotesize -} & {\footnotesize -} & {\footnotesize -} & {\footnotesize -} & {\footnotesize 58.0} & {\footnotesize 20.9}\tabularnewline
{\footnotesize SG-Nav \cite{SG-Nav-nips24}} & {\footnotesize$\checkmark$} & {\footnotesize 40.2} & {\footnotesize 16.0} & {\footnotesize 54.0} & {\footnotesize 24.9} & {\footnotesize 49.6} & {\footnotesize 25.5}\tabularnewline
{\footnotesize UniGoal \cite{unigoal_cvpr25}} & {\footnotesize$\checkmark$} & {\footnotesize 41.0} & {\footnotesize 16.4} & {\footnotesize 54.5} & {\footnotesize 25.1} & {\footnotesize -} & {\footnotesize -}\tabularnewline
{\footnotesize ApexNAV \cite{ApexNAV_RAL25}} & {\footnotesize$\checkmark$} & {\footnotesize 39.2} & {\footnotesize 17.8} & {\footnotesize 59.6} & {\footnotesize 33.0} & {\footnotesize 76.2} & {\footnotesize 38.0}\tabularnewline
{\footnotesize BeliefMapNav \cite{beliefmapnav}} & {\footnotesize$\checkmark$} & {\footnotesize 37.3} & {\footnotesize 17.6} & {\footnotesize 61.4} & {\footnotesize 30.6} & {\footnotesize -} & {\footnotesize -}\tabularnewline
\hline 
{\footnotesize TrajRAG (Ours)} & {\footnotesize$\checkmark$} & {\footnotesize \textbf{42.6}} & {\footnotesize \textbf{18.0}} & {\footnotesize \textbf{62.5}} & {\footnotesize \textbf{33.9}} & {\footnotesize \textbf{78.1}} & {\footnotesize \textbf{40.2}}\tabularnewline
\hline 
\end{tabular}
\vspace{-10pt}
\end{table}

\textbf{Zero-shot ObjectNav}. 
We compare our method with prior works on three challenging datasets in Tab.~\ref{tab:ObjectNav}, where competing methods are categorized by their support for open-vocabulary (OV) object goals.
Our proposed TrajRAG establishes new state-of-the-art performance across all benchmarks. It achieves the highest Success Rate (SR) and Success weighted by Path Length (SPL) on the MP3D, HM3Dv1, and HM3Dv2.

TrajRAG’s superior performance stems from its core innovation of leveraging historical navigation experiences. 
By retrieving the most relevant trajectories from an external knowledge base that accounts for both scene context and navigational relevance, the agent can make more informed decisions.
This retrieval-augmented strategy guides agent's exploration and long-horizon planning, helping it avoid myopic behavior and navigate more efficiently towards open-vocabulary goals in complex, unseen environments.


\section{Conclusion}

In this paper, we introduced Trajectory RAG (TrajRAG), a retrieval-augmented generation framework for zero-shot ObjectNav that enhances navigation reasoning by retrieving geometric–semantic experiences. 
To ensure compact and efficient storage, we propose a topological-polar (topo-polar) trajectory representation that enables self-checking by detecting and pruning redundant segments, while supporting duplicate detection during trajectory integration.
Furthermore, TrajRAG employs a hierarchical chunking architecture to enable efficient retrieval of relevant experiences. Experiments on MP3D, HM3D-v1, and HM3D-v2 demonstrate that TrajRAG effectively retrieves geometric–semantic experiences and improves navigation performance in zero-shot ObjectNav.

\section*{Acknowledgements}

This work was supported in part by the National Natural Science Foundation of China under Grant 62125207, Grant 62495084, Grant 62272443, and Grant U23B2012, in part by the Beijing Natural Science Foundation under Grant L242020, in part by the Postdoctoral Fellowship Program and China Postdoctoral Science Foundation under Grant Number BX20250391, and in part by the Suzhou Science and Technology Plan Project under grant SYG2024082.

{
    \small
    \bibliographystyle{ieeenat_fullname}
    \bibliography{main}
}


\end{document}